\documentclass[10pt,twocolumn, letterpaper]{article} 

\newcommand{\final}{1} 
\newcommand{\forReview}{0} 

\usepackage{color}
\usepackage{ifthen}
\usepackage{float}
\usepackage{alltt}
\usepackage{newlfont} 
\usepackage{wrapfig}
\usepackage{booktabs}
\usepackage{multirow}
\usepackage{comment}
\usepackage{gensymb}

\usepackage{comment}
\usepackage{amsmath}
\usepackage{amssymb}
\usepackage{amsthm}
\usepackage{cvpr}
\usepackage{times}
\usepackage{epsfig}
\usepackage{graphicx}
\usepackage{amsmath}
\usepackage{amssymb}
\usepackage{amsthm}
\usepackage{subcaption}
\usepackage{microtype}

\definecolor{DeltaColor}{rgb}{0.039,0.73,0.71}
\definecolor{SetaColor}{rgb}{0.867, 0.0235, 0.376}
\definecolor{SigmaColor}{rgb}{0.98,0.45,0.0}
\definecolor{HaoColor}{rgb}{0.8,0,0}
\definecolor{AlphaColor}{rgb}{0,0,0.8}
\definecolor{BetaColor}{rgb}{0.8,0,0.8}
\definecolor{GammaColor}{rgb}{0.5,0,0.7}
\definecolor{EpsilonColor}{rgb}{0.353,0.725,0.906}
\definecolor{TauColor}{rgb}{0.423,0.235,0.192}
\newcommand{\ryota}[1]{{\color{SetaColor} Ryota: #1 $\qed$}}
\newcommand{\shunsuke}[1]{{\color{AlphaColor} Shunsuke: #1 $\qed$}}
\newcommand{\zeng}[1]{{\color{SigmaColor} Zeng: #1 $\qed$}}
\newcommand{\hao}[1]{{\color{GammaColor} Hao: #1 $\qed$}}
\newcommand{\chongyang}[1]{{\color{DeltaColor} Chongyang: #1 $\qed$}}
\newcommand{\weikai}[1]{{\color{EpsilonColor} Weikai: #1 $\qed$}}

\newcommand{\warning}[1]{{\it\color{red} #1}}
\newcommand{\note}[1]{{\it\color{blue} #1}}
\newcommand{\nothing}[1]{}

\definecolor{AudioColor}{rgb}{0.56,0.34,0.62}

\definecolor{DeadlineColor}{rgb}{0.9,0.4,0} 
\newcommand{\deadline}[1]{{\bf\color{DeadlineColor} ETA: #1}}

\definecolor{figred}{rgb}{1,0,0}
\definecolor{figgreen}{rgb}{0,0.6,0}
\definecolor{figblue}{rgb}{0,0,1}
\definecolor{figpink}{rgb}{1,0.63,0.63}

\ifthenelse{\equal{\final}{1}}
{
\renewcommand{\l}[1]{}
\renewcommand{\zeng}[1]{}
\renewcommand{\shunsuke}[1]{}
\renewcommand{\ryota}[1]{}
\renewcommand{\hao}[1]{}
\renewcommand{\chongyang}[1]{}
\renewcommand{\weikai}[1]{}
\renewcommand{\warning}[1]{}
\renewcommand{\note}[1]{}
\renewcommand{\deadline}[1]{}
}
{}

\newcounter{pccount}
\floatstyle{plain}

\newcommand{\filename}[1]{\url{#1}}
\newcommand{\foldername}[1]{\url{#1}}

\newcommand{\loss}{\mathcal{L}}
\newcommand{\bceloss}{\loss_{BCE}}
\newcommand{\adversarialloss}{\loss_{adv}}
\newcommand{\silhouette}{\mathbf{S}}
\newcommand{\silnet}{\mathcal{G}_{s}}
\newcommand{\pose}{\mathbf{P}}
\newcommand{\sourcepose}{\pose_{s}}
\newcommand{\targetpose}{\pose_{t}}
\newcommand{\sourcesil}{\silhouette_{s}}
\newcommand{\targetsil}{\silhouette_{t}}

\newcommand{\vh}{\mathcal{H}}
\newcommand{\view}{\mathcal{V}}
\newcommand{\bin}{\mathcal{B}}

\newcommand{\fronttoback}{\mathcal{G}_{t}}
\newcommand{\featurematchloss}{\loss_{FM}}
\newcommand{\vggloss}{\loss_{VGG}}
\newcommand{\inputimage}{\mathbf{I}}
\newcommand{\frontimage}{\inputimage_{f}}

\newcommand{\frontsil}{\silhouette_{f}}
\newcommand{\fakebackimage}{\hat{\inputimage}_{b}}

\newcommand{\normal}{\mathbf{n}}
\newcommand{\camera}{\mathbf{c}}

\ifthenelse{\equal{\forReview}{1}}
{
\usepackage[pagebackref=true,breaklinks=true,letterpaper=true,colorlinks,bookmarks=false]{hyperref}
\ifcvprfinal\pagestyle{empty}\fi
}
{
\usepackage[breaklinks=true,colorlinks,bookmarks=false]{hyperref}
\cvprfinalcopy 

\newcommand*\samethanks[1][\value{footnote}]{\footnotemark[#1]}

\author{Ryota Natsume\textsuperscript{1,3}\thanks{Joint first authors} \hspace{0.3in} Shunsuke Saito\textsuperscript{1,2}\samethanks \hspace{0.3in} Zeng Huang\textsuperscript{1,2} \hspace{0.3in} Weikai Chen\textsuperscript{1} \\
Chongyang Ma\textsuperscript{4} \hspace{0.3in} Hao Li\textsuperscript{1,2,5} \hspace{0.3in} Shigeo Morishima\textsuperscript{3}
\vspace{5pt}
\\
\textsuperscript{1}{USC Institute for Creative Technologies} \hspace{0.3in} \textsuperscript{2}{University of Southern California} \\ \textsuperscript{3}{Waseda University} \hspace{0.3in} \textsuperscript{4}{Snap Inc.} \hspace{0.3in} \textsuperscript{5}{Pinscreen}
}
}
\def\cvprPaperID{1456} 

\graphicspath{
{figures}
{images}
}

\begin{document}

\title{SiCloPe: Silhouette-Based Clothed People}

\maketitle

\begin{abstract}{~}
We introduce a new silhouette-based representation for modeling clothed human bodies using deep generative models. Our method can reconstruct a complete and textured 3D model of a person wearing clothes from a single input picture. Inspired by the visual hull algorithm, our implicit representation uses 2D silhouettes and 3D joints of a body pose to describe the immense shape complexity and variations of clothed people. Given a segmented 2D silhouette of a person and its inferred 3D joints from the input picture, we first synthesize consistent silhouettes from novel view points around the subject. The synthesized silhouettes which are the most consistent with the input segmentation are fed into a deep visual hull algorithm for robust 3D shape prediction. 
We then infer the texture of the subject's back view using the frontal image and segmentation mask as input to a conditional generative adversarial network.
Our experiments demonstrate that our silhouette-based model is an effective representation and the appearance of the back view can be predicted reliably using an image-to-image translation network. While classic methods based on parametric models often fail for single-view images of subjects with challenging clothing, our approach can still produce successful results, which are comparable to those obtained from multi-view input.
\end{abstract}

\section{Introduction}
\label{sec:intro}


The ability to digitize and predict a complete and fully textured 3D model of a clothed subject from a single view can open the door to endless applications, ranging from virtual and augmented reality, gaming, virtual try-on, to 3D printing. A system that could generate a full-body 3D avatar of a person by simply taking a picture as input would significantly impact the scalability of producing virtual humans for immersive content creation, as well as its attainability by the general population. Such single-view inference is extremely difficult due to the vast range of possible shapes and appearances that clothed human bodies can take in natural conditions. Furthermore, only a 2D projection of the real world is available and the entire back view of the subject is missing.

\begin{figure}[t]
\begin{center}
  \includegraphics[width=\linewidth]{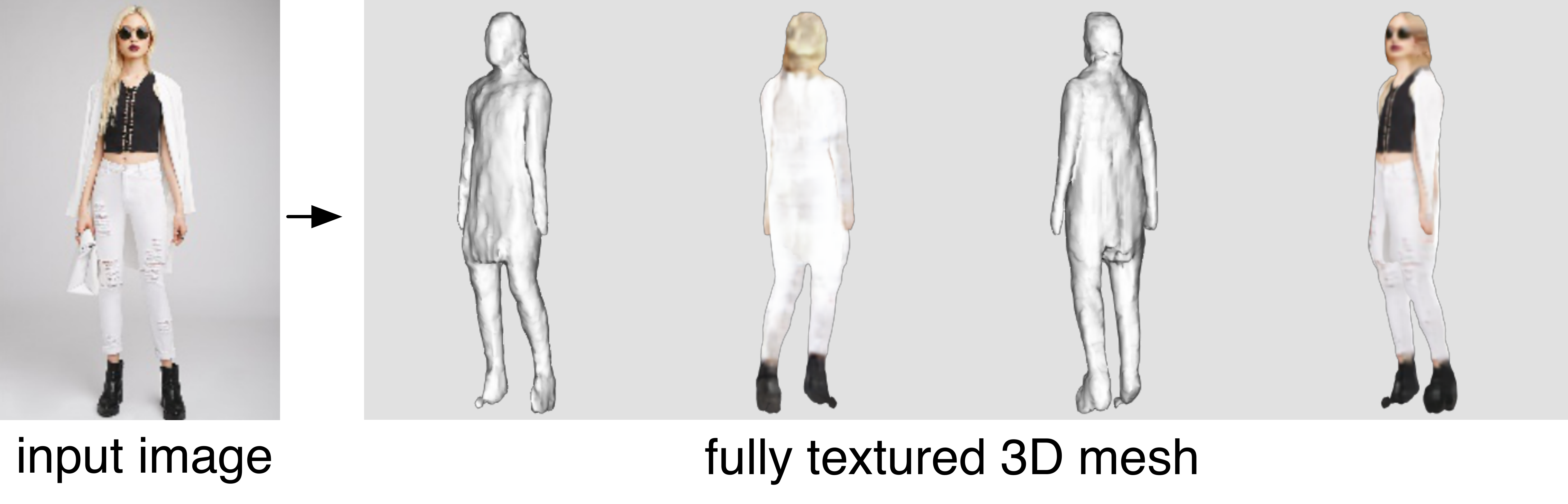}
\end{center}
\vspace{-10pt} 
\caption{Given a single image of a person from the frontal view, we can automatically reconstruct a complete and textured 3D clothed body shape.}
\label{fig:teaser}
\end{figure}


While 3D range sensing~\cite{li2013three,NewcombeFS15} and photogrammetry~\cite{Seitz:2006} are popular ways of obtaining complete 3D models, they are restricted to a tedious scanning process or require specialized equipment.
The modeling of humans from a single view, on the other hand, has been facilitated by the availability of large 3D human model repositories~\cite{anguelov2005scape,loper2015smpl}, where a {\em parametric model} of human shapes is used to guide the reconstruction process~\cite{bogo2016keep}. 
However, these parametric models only represent naked bodies and do not describe the clothing geometry nor the texture.
Another option is to use a {\em pre-captured template} of the subject in order to handle new poses~\cite{Xu:2018:MHP}, but such an approach is limited to the recording of one particular person.


In this work, we propose a deep learning based non-parametric approach for generating the geometry and texture of clothed 3D human bodies from a single frontal-view image.
Our method can predict fine-level geometric details of clothes and generalizes well to new subjects different from those being used during training (See Figure~\ref{fig:teaser}).


While directly estimating 3D volumetric geometry from a single view is notoriously challenging and likely to require a large amount of training data as well as extensive parameter tuning, two cutting-edge deep learning techniques have shown that impressive results can be obtained using 2D silhouettes from very sparse views~\cite{Huang18ECCV,varol18_bodynet}. Inspired by these approaches based on visual hull, we propose the first algorithm to predict 2D silhouettes of the subject from multiple views given an input segmentation, which implicitly encodes 3D body shapes. We also show that a sparse 3D pose estimated from the 2D input~\cite{bogo2016keep,Rogez:2018:LCR-Net} can help reduce the dimensionality of the shape deformation and guide the synthesis of consistent silhouettes from novel views. 

We then reconstruct the final 3D geometry from multiple silhouettes using a deep learning based visual hull technique by incorporating a clothed human shape prior. Since silhouettes from arbitrary views can be generated, we further improve the reconstruction result by greedily choosing view points that will lead to improved silhouette consistency.  
To fully texture the reconstructed geometry, we propose to train an image-to-image translation framework to infer the color texture of the back view given the input image from the frontal view.



We demonstrate the effectiveness of our method on a variety of input data, including both synthetic and real ones.
We also evaluate major design decisions using ablation studies and compare our approach with state of the art single-view as well as multi-view reconstruction techniques.

In summary, our contributions include:
\begin{itemize}
\setlength\itemsep{0em} 
\item The first non-parametric solution for reconstructing fully textured and clothed 3D humans from a single-view input image.
\item An effective two-stage 3D shape reconstruction pipeline that consists of predicting multi-view 2D silhouettes from a single input segmentation and a novel deep visual hull based mesh reconstruction technique with view sampling optimization.
\item An image-to-image translation framework to reconstruct the texture of a full body from a single photo.
\end{itemize}

\section{Related Work}
\label{sec:related_work}

\paragraph{Multi-view reconstruction.}
Due to the geometric complexity introduced by garment deformation and self occlusions, reconstructing clothed human bodies usually requires images captured from multiple viewpoints.
Early attempts in this direction have extensively explored visual hull based approaches~\cite{matusik2000image, vlasic2008articulated, furukawa2006carved, esteban2004silhouette, cheung2003visual, franco2006visual} due to its efficiency and robustness to approximate the underlying 3D geometry.
However, a visual hull based representation cannot handle concave regions nor generate good approximations of fine-scale details especially when the number of input views is limited.
To address this issue, detailed geometry are often captured using techniques based on multi-view stereo constraints~\cite{starck2007surface, zitnick2004high, waschbusch2005scalable, Seitz:2006,vlasic2009dynamic,furukawa2010accurate,wu2011shading}. A number of techniques~\cite{yang2016estimation,pons2017clothcap,zhang-CVPR17} exploit motion cues as additional priors for a more accurate digitization of body shapes.

Some more recent research have focused on monocular input capture, with the goal of making human modeling more accessible to end users
~\cite{Xu:2018:MHP,alldieck2018video,alldieck2018detailed}.
With the recent advancement of deep learning, an active research direction is to encode shape prior in a deep neural network in order to model the complexity of human body and garment deformations. To this end, Huang~\etal~\cite{Huang18ECCV} and Gilbert~\etal~\cite{gilbert:eccv:2018} have presented techniques that can synthesize clothed humans in a volumetric form from highly sparse views.
Although the number of input views are reduced, both methods still require a carefully calibrated capture system.
In this work, we push the envelop by reducing the input to a single unconstrained input photograph. 




\paragraph{Single-view reconstruction.}
To reduce the immense solution space of human body shapes, several 3D body model repositories, e.g. SCAPE~\cite{anguelov2005scape} and SMPL~\cite{loper2015smpl}, have been introduced, which have made the single-view reconstruction of human bodies more tractable.
In particular, a 3D parametric model is built from such database, which uses pose and shape parameters of the 3D body to best match an input image~\cite{balan2007detailed,guan2009estimating,bogo2016keep, lassner2017unite}.
As the mapping between the body geometry and the parameters of the deformable model is highly non-linear, alternative approaches based on deep learning have become increasingly popular.
The seminal work of Dibra~\etal~\cite{dibra2016hs, dibra2017human} introduces deep neural networks to estimate the shape parameters from a single input silhouette.
More recent works predict body parameters of the popular SMPL model~\cite{bogo2016keep} by either minimizing the silhouette matching error~\cite{tan2017indirect}, joint error based on the silhouette and 2D joints~\cite{tung2017self}, or an adversarial loss that can distinguish unrealistic reconstruction output~\cite{hmrKanazawa17}.
Concurrent to our work, Weng~\etal~\cite{weng2018photo} present a method to animate a person in 3D from a single image based on the SMPL model and 2D warping.

Although deformable models offer a low-dimensional embedding of complex non-rigid human body shapes, they are not suitable for modeling of fine-scale clothing details.
To address this issue, additional information such as 2D~\cite{wei2016cpm, cao2017realtime} and 3D body pose~\cite{mehta2017vnect, yang20183d, guler2018densepose} has been incorporated to help recover clothed body geometry without relying on a template mesh.
BodyNet~\cite{varol18_bodynet} for instance, estimates volumetric body shapes from a single image based on an intermediate inference of 2D pose, 2D part segmentation, as well as 3D pose. 
The latest advances in novel view synthesis of human pose~\cite{ma2017pose, balakrishnan2018synthesizing} and 3D shape~\cite{zhu2018view, zhou2016view, rematas2017novel} have demonstrated the ability of obtaining multi-view inference from a single image.
In this work, we introduce an approach that combines 3D poses estimation with the inference of silhouettes from novel views for predicting high-fidelity clothed 3D human shapes from a single photograph. We show that our method can achieve reasonably accurate reconstructions automatically without any template model. 






\nothing{
\paragraph{Deep learning for 3D shapes.}
As more and more 3D shape datasets become available~\cite{Chang:2015:ShapeNet,Dai:2017:ScanNet\nothing{,Sun:2018:Pix3D}}, image based shape reconstruction has emerged as a popular research topic in the computer vision community.
A number of recent techniques in this domain can predict 3D output from a single input image based on different shape representations, including uniform voxels~\cite{qi2016volumetric,jackson2017large,Saito:2018:3HS,varol18_bodynet}\nothing{\cite{Tulsiani:2017:MVS,Wu:2017:MarrNet,Wu:2018:ShapeHD,Yang:2018:LSV,Henderson:2018:LGR}}, octree~\cite{Tatarchenko:2017:OGN,Wang:2017:OCNN,Wang:2018:AOCNN}, point cloud~\cite{qi2016pointnet, qi2017pointnet++, zhou2017voxelnet}\nothing{\cite{Fan:2017:PSG,Xia:2018:RealPoint3D}}, depth and surface normals (2.5D sketches)~\cite{Wiles:2018:SRP,Li:2018:SVBRDF}, layered structures~\cite{Tulsiani:2018:LSI,Richter:2018:MBP}, as well as surface meshes~\cite{kato2018neural,cmrKanazawa18,Groueix:2018:AtlasNet,Wang:2018:Pixel2Mesh,Pumarola:2018:GAN}.
}



\begin{figure*}[t]
\centering
\includegraphics[width=\linewidth]{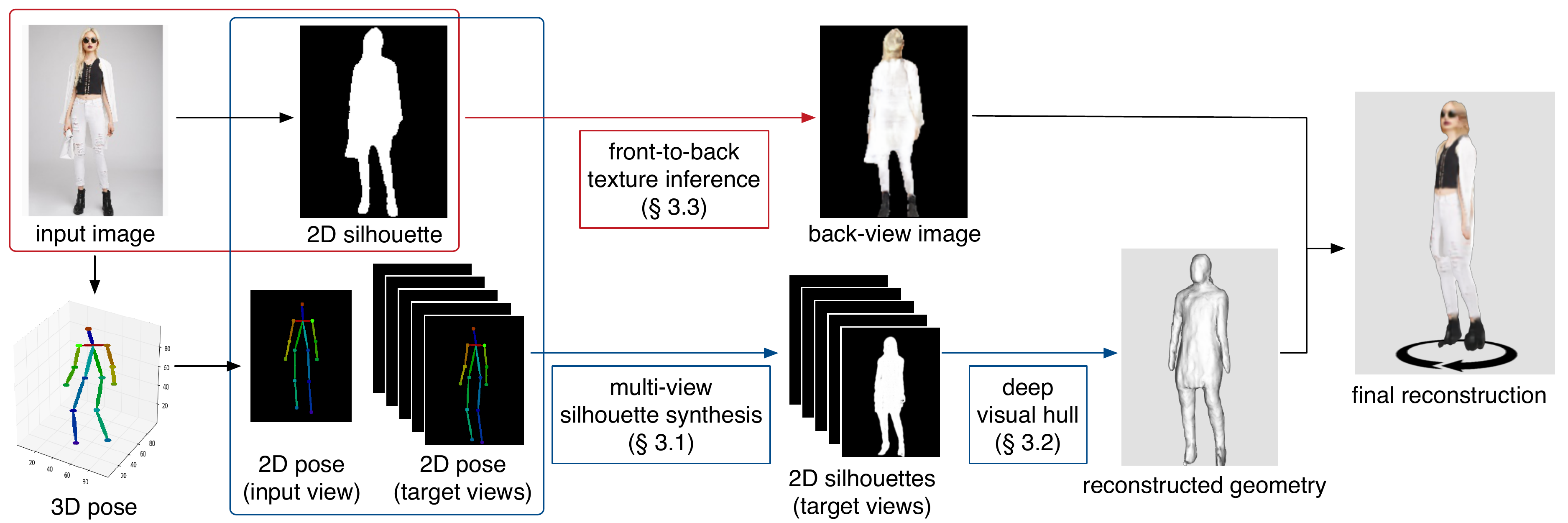}
\vspace{-5pt}
\caption{Overview of our framework.}
\label{fig:overview}
\end{figure*}

\section{Method}
\label{sec:method}

Our goal is to reconstruct a wide range of 3D clothed human body shapes with a complete texture from a single image of a person in frontal view. Figure~\ref{fig:overview} illustrates an overview of our system.
Given an input image, we first extract the 2D silhouette and 3D joint locations, which are fed into a silhouette synthesis network to generate plausible 2D silhouettes from novel viewpoints (Sec.~\ref{sec:silnet}). The network produces multiple silhouettes with known camera projections, which are used as input for 3D reconstruction via visual hull algorithms \cite{vlasic2008articulated}. However, due to possible inconsistency between the synthesized silhouettes, the subtraction operation of visual hull tends to excessively erode the reconstructed mesh. To further improve the output quality, we adopt a deep visual hull algorithm similar to Huang~\etal~\cite{Huang18ECCV} with a greedy view sampling strategy so that the reconstruction results account for domain-specific clothed human body priors (Sec.~\ref{sec:dvhull}). Finally, we inpaint the non-visible body texture on the reconstructed mesh by inferring the back view of the input image using an image-to-image translation network (Sec.~\ref{sec:f2b}).

\subsection{Multi-View Silhouette Synthesis}
\label{sec:silnet}

\begin{figure}[t]
\centering
\includegraphics[width=\linewidth]{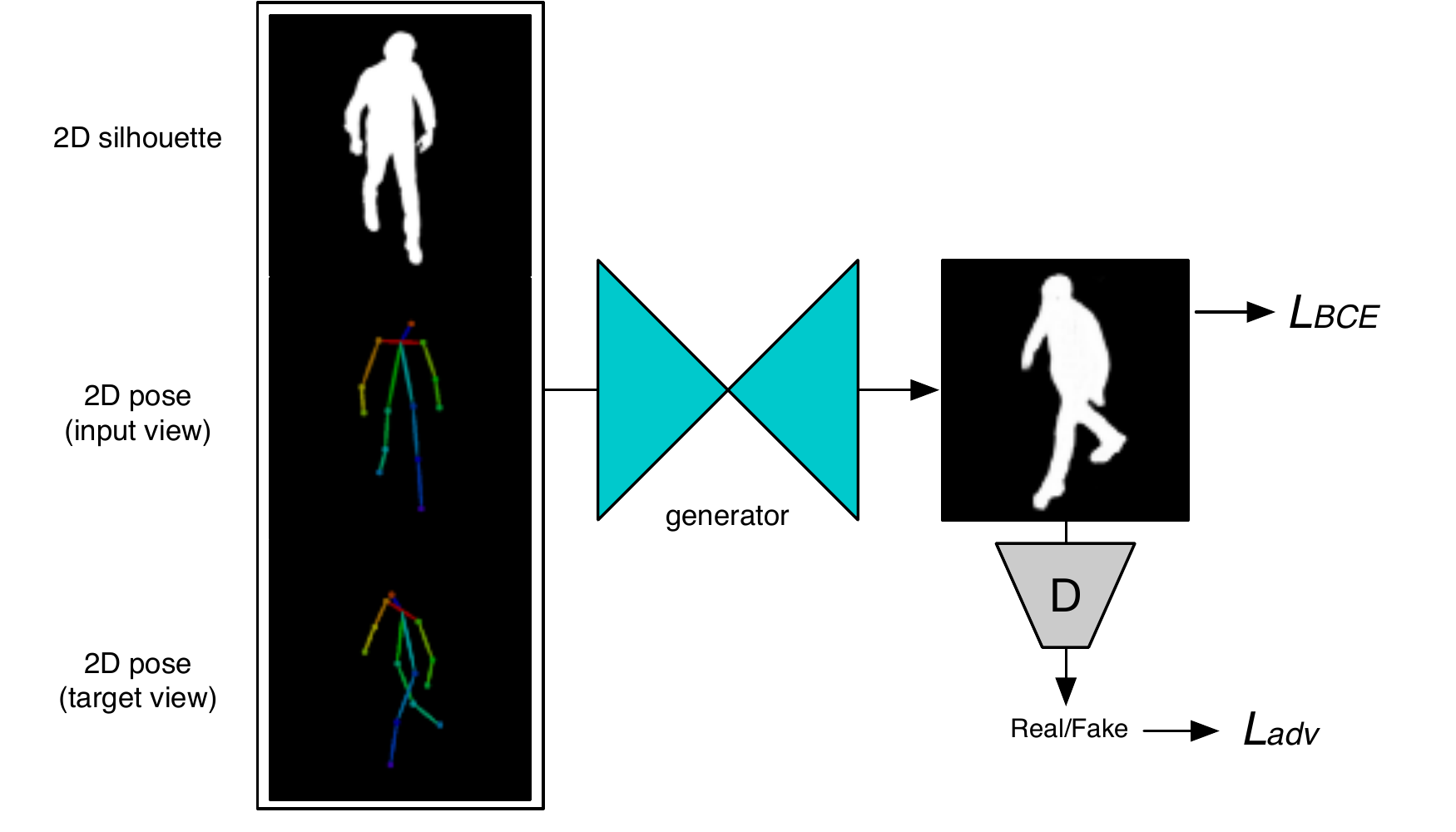}
\vspace{-15pt}
\caption{Illustration of our silhouette synthesis network.}
\label{fig:silnet_overview}
\end{figure}

We seek an effective human shape representation that can handle the shape complexity due to different clothing types and deformations.
Inspired by visual hull algorithms~\cite{matusik2000image} and recent advances in conditional image generation~\cite{esser2018variational,ma2018disentangled,zhang2017image,zhang2018densely,li2018context}, we propose to train a generative network for synthesizing 2D silhouettes from viewpoints other than the input image (see Figure~\ref{fig:silnet_overview}).
We use these silhouettes as an intermediate implicit representation for the 3D shape inference.

Specifically, given the subject's 3D pose, estimated from the input image as a set of 3D joint locations, we project the 3D pose onto the input image and a target image plane to get the 2D pose $\sourcepose$ in the source view and the pose $\targetpose$ in the target view, respectively.
Our silhouette synthesis network $\silnet$ takes the input silhouette $\sourcesil$ together with $\sourcepose$ and $\targetpose$ as input, and predicts the 2D silhouette in the target view $\targetpose$:
\begin{equation}
\begin{split}
\targetsil = \silnet \big(\sourcesil, \sourcepose, \targetpose \big).
\label{eqn:silnet_synthesis}
\end{split}
\end{equation}

\begin{figure}[t]
\begin{center}
  \includegraphics[width=\linewidth, trim=0px 0px 0px 30px, clip]{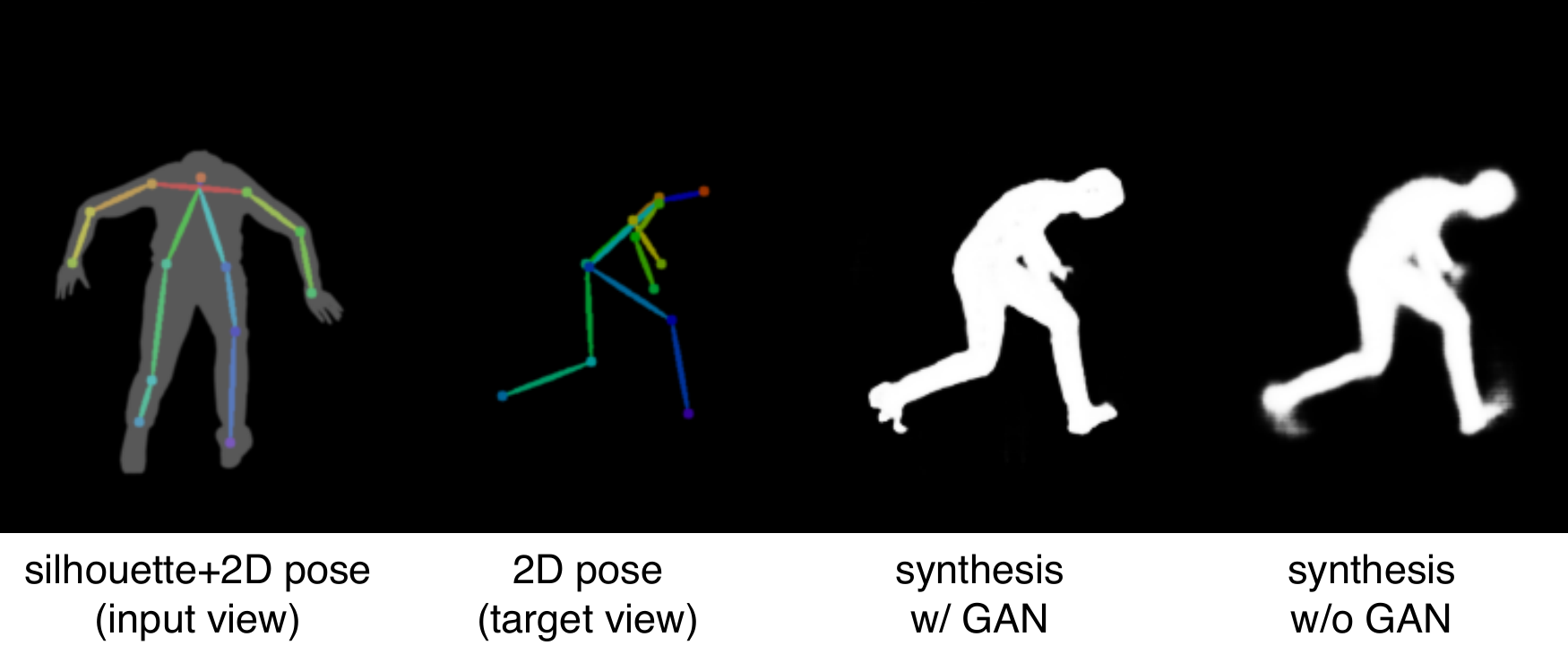}
\end{center}
\vspace{-15pt} 
\caption{GAN helps generate clean silhouettes in presence of ambiguity in silhouette synthesis from a single view.}
\label{fig:silnet_gan}
\end{figure}

Our loss function for training the network $\silnet$ consists of reconstruction errors of the inferred silhouettes using a binary cross entropy loss $\bceloss$ and a patch-based adversarial loss $\adversarialloss$~\cite{pix2pix2017}. The total objective function is given by
\begin{equation}
\begin{split}
\loss = \lambda_{BCE} \cdot \bceloss + \adversarialloss,
\label{eqn:silnet_loss}
\end{split}
\end{equation}
where the relative weight $\lambda_{BCE}$ is set to $750$.
In particular, the adversarial loss turns out to be critical for synthesizing sharp and detailed silhouettes.
Figure~\ref{fig:silnet_gan} shows that the loss function with the adversarial term generate much sharper silhouettes, while without an adversarial loss would lead to blurry synthesis output.

\paragraph{Discussions.}
The advantages of using silhouettes to guide the 3D reconstruction are two-fold. First, since silhouettes are binary masks, the synthesis can be formulated as a pixel-wise classification problem, which can be trained more robustly without the need of complex loss functions or extensive hyper parameter tuning in contrast to novel view image synthesis~\cite{ma2017pose, balakrishnan2018synthesizing}. Second, the network can predict much a higher spatial resolution since it does not store 3D voxel information explicitly, as with volumetric representations~\cite{varol18_bodynet}, which are bounded by the limited output resolution.

\nothing{
\begin{itemize}
    \item given silhouettes and 2d joint locations in the input and the target view, the network synthesizes the silhouette from the target view.
    \item explain how we project 3d joints onto each view. Note that 2d projected joints are encoded as RGB image
    \item mention that silhouette synthesis is pixel-wise classification problem, which can be more stably trained without bells and whistles in contrast to novel view image synthesis problems.
    \item loss: BCE ($w=750$) + (unconditional)GAN($w=1.0$)
    \item explain why GAN is necessary (dealing with ambiguity, multi-modal nature of target)
\end{itemize}
}

\subsection{Deep Visual Hull Prediction}
\label{sec:dvhull}


Although our silhouette synthesis algorithm generates sharp prediction of novel-view silhouettes, the estimated results may not be perfectly consistent as the conditioned 3D joints may fail to fully disambiguate the details in the corresponding silhouettes (e.g., fingers, wrinkles of garments).
Therefore, naively applying conventional visual hull algorithms is prone to excessive erosion in the reconstruction, since the visual hull is designed to subtract the inconsistent silhouettes in each view.   
To address this issue, we propose a deep visual hull network that reconstructs a plausible 3D shape of clothed body without requiring perfectly view-consistent silhouettes by leveraging the shape prior of clothed human bodies. 

In particular, we use a network structure based on \cite{Huang18ECCV}.
At a high level, Huang~\etal~\cite{Huang18ECCV} propose to map 2D images to a 3D volumetric field through a multi-view convolutional neural network.
The 3D field encodes the probabilistic distribution of 3D points on the captured surface.
By querying the resulting field, one can instantiate the geometry of clothed human body at an arbitrary resolution.
However, unlike their approach which takes carefully calibrated color images from fixed views as input, our network only consumes the probability maps of novel-view silhouettes, which can be inconsistent across different views.
Although arbitrary number of novel-view silhouettes can be generated, it remains challenging to properly select optimal input views to maximize the network performance. Therefore, we introduce several improvements to increase the reconstruction accuracy.


\paragraph{Greedy view sampling.}
We propose a greedy view sampling strategy to choose proper views that can lead to better reconstruction quality.
Our key idea is to generate a pool of candidate silhouettes and then select the views that are most consistent in a greedy manner.
In particular, the candidate silhouettes are rendered from $12$ view bins $\{\bin_i\}$: the main orientations of the bins are obtained by uniformly sampling $12$ angles in the yaw axis.
The first bin only contains the input view and thus has to be aligned with the orientation of the input viewpoint.
Each of the other bins consists of $5$ candidate viewpoints, which are distributed along the pitch axis with angles sampled from  $\{0^{\circ}, 15^{\circ}, 30^{\circ}, 45^{\circ}, 60^{\circ} \}$.
In the end, we obtain $55$ candidate viewpoints $\{\view_i\}$ to cover most parts of the 3D body.

To select the views with maximal consistency, we first compute an initial bounding volume of the target model based on the input 3D joints.
We then carve the bounding volume using the silhouette of the input image and obtain a coarse visual hull $\vh_1$.
The bins with remaining views are iterated in a clockwise order, i.e., only one candidate view will be sampled from each bin at the end of the sampling process.
Starting from the second bin $\bin_2$, the previously computed visual hull $\vh_1$ is projected to its enclosed views.
The candidate silhouette that has the maximum 2D intersection over union (IoU) with $\vh_1$'s projection will be selected as the next input silhouette for our deep visual hull algorithm.
After the best silhouette $\hat{\view}_2$ is sampled from $\bin_2$, $\vh_1$ is further carved by $\hat{\view}_2$ and the updated visual hull $\vh_2$ is passed to the next iteration.
We iterated until all the view bins have been sampled.

The selected input silhouettes generated by our greedy view sampling algorithm are then fed into a deep visual hull network.
The choice of our network design is similar to that of \cite{Huang18ECCV}.
The main difference lies in the format of inputs.
Specifically, in addition to multi-view silhouettes, our network also takes the 2D projection of the 3D pose as additional channel concatenated with the corresponding silhouette.
This change helps to regularize the body part generation by passing the semantic supervision to the network and thus improves robustness.
Moreover, we also reduce some layers of the network of \cite{Huang18ECCV} to achieve a more compact model and to prevent overfitting.
The detailed architecture is provided in our supplementary materials.


\nothing{
\shunsuke{
\begin{itemize}
    \item briefly summarize [Huang et al.]~\cite{Huang18ECCV}
    \item we takes the inferred silhouettes instead of images
    \item mention that synthesized silhouettes might be inconsistent, which results in excessively eroded reconstruction.
    \item therefore we propose to optimize sampling views so that it maximizes the consistency with the input view while minimizing the overall volume
    \item illustrates that it produces better reconstruction results qualitatively and quantitatively.
\end{itemize}
}
}

\subsection{Front-to-Back Texture Synthesis}
\label{sec:f2b}

\begin{figure}[t]
\centering
\includegraphics[width=\linewidth]{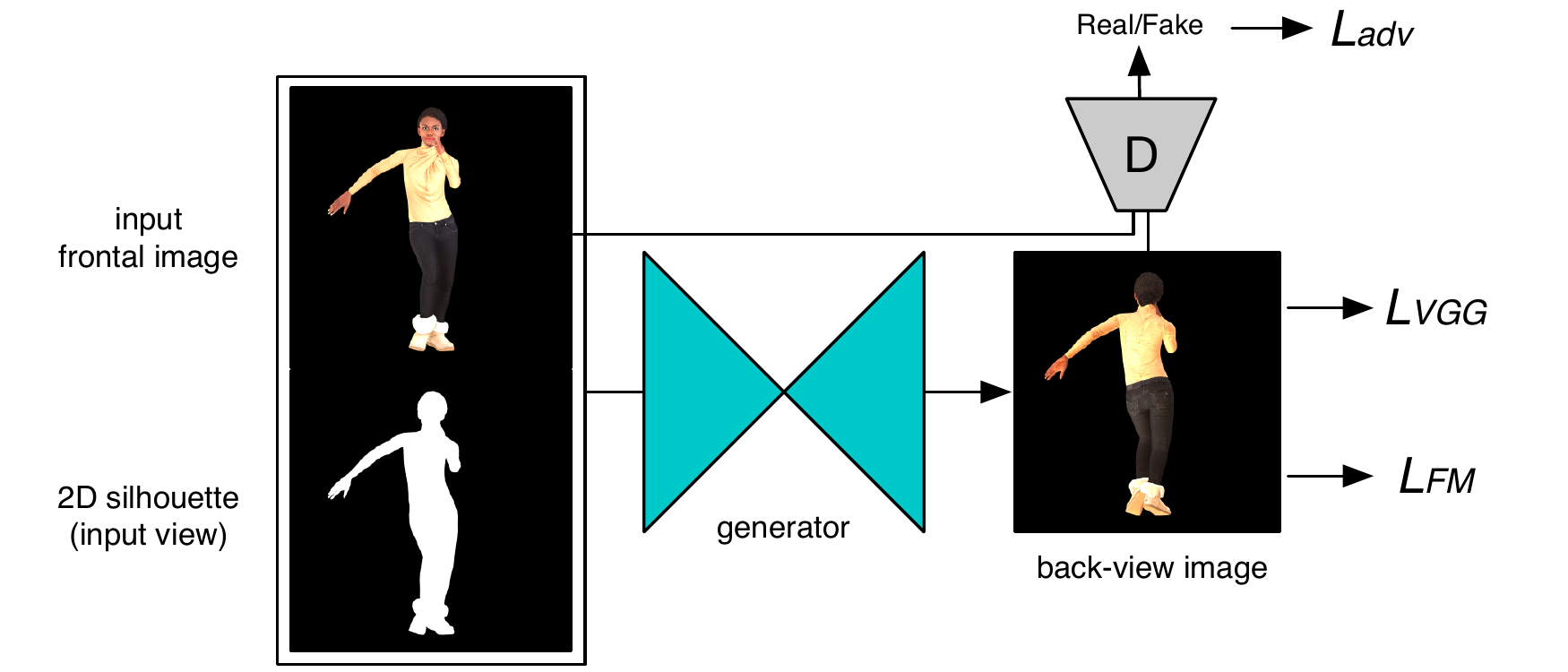}
\vspace{-15pt} 
\caption{Illustration of our front-to-back synthesis network.}
\label{fig:f2b_overview}
\end{figure}

When capturing the subject from a single viewpoint, only one side of the texture is visible and therefore predicting the other side of the texture appearance is required to reconstruct a fully textured 3D body shape.
Our key observation is that the frontal view and the back view of a person are spatially aligned by sharing the same contour and many visual features.
This fact has inspired us to solve the problem of back-view texture prediction using an {\em image-to-image} translation framework based conditional generative adversarial network.
Specifically, we train a generator $\fronttoback$ to predict the back-view texture $\fakebackimage$ from the frontal-view input image $\frontimage$ and the corresponding silhouette $\frontsil$:
\begin{equation}
\begin{split}
\fakebackimage = \fronttoback(\frontimage, \frontsil).
\label{eqn:backview_synthesis}
\end{split}
\end{equation}

We train the generator $\fronttoback$ in a supervised manner by leveraging textured 3D human shape repositories to generate a dataset that suffices for our training objective (Sec.~\ref{sec:impl}).
Adopted from a high-resolution image-to-image translation network~\cite{wang2018pix2pixHD}, our loss function consists of a feature matching loss $\featurematchloss$ that minimizes the discrepancy of intermediate layer activation of the discriminator $D$, a perceptual loss $\vggloss$ using a VGG19 model pre-trained for image classification task~\cite{simonyan2014very}, and an adversarial loss $\adversarialloss$ conditioned by the input frontal image (see Figure~\ref{fig:f2b_overview}). The total objective is defined as:
\begin{equation}
\begin{split}
\loss = \lambda_{FM} \cdot \featurematchloss + \lambda_{VGG} \cdot \vggloss + \adversarialloss,
\label{eqn:f2b_loss}
\end{split}
\end{equation}
where we set the relative weights as $\lambda_{FM} = \lambda_{VGG} = 10.0$ in our experiments.

The resulting back-view image is used to complete the per-vertex color texture of the reconstructed 3D mesh. If the dot product between the surface normal $\normal$ in the input camera space and the camera ray $\camera$ is negative (i.e., surface is facing towards the camera), the vertex color is sampled from the input view image at the corresponding screen coordinate. Likewise, if the dot product is positive (i.e., surface is facing in the opposite direction), the vertex color is sampled from the synthesized back-view image. When the surface is perpendicular to the camera ray (i.e., $|\normal \cdot \camera| \leq \epsilon = 1.0\times10^{-4}$), we blend the colors from the front and back views so that there are no visible seams across the boundary.

\nothing{
\begin{itemize}
    \item this properly allows a network with skip connections to aggregate local information and global context better.
    \item front2back conditional gan (pair of front and back to check if it's real or fake)
    \item Loss: Feature matching($w=10.0$), VGG loss($w=10.0$), f2b-GAN($w=1.0$)
    \item FM loss and VGG loss are following Pixe2PixHD~\cite{wang2018pix2pixHD}
    \item explain how we stitch texture (simple feathering)
\end{itemize}
}

\subsection{Implementation Details}
\label{sec:impl}

\paragraph{Body mesh datasets.}
We have collected $73$ rigged meshes with full textures from aXYZ\footnote{\url{https://secure.axyz-design.com/}}\nothing{~\cite{axyz}} and $194$ meshes from Renderpeople\footnote{\url{https://renderpeople.com/3d-people/}}.
We randomly split the dataset into a training set and a test set of $247$ and $20$ meshes, respectively.
We apply $48$ animation sequences (such as walking, waving, and Samba dancing) from Mixamo\footnote{\url{https://www.mixamo.com/}} to each mesh from Renderpeople to collect body meshes of different poses.
Similarly, the meshes from aXYZ have been animated into $11$ different sequences.
To render synthetic training data, we have also obtained $163$ second-order spherical harmonics of indoor environment maps from HDRI Haven\footnote{\url{https://hdrihaven.com/}} and they are randomly rotated around the yaw axis.

\paragraph{Camera settings for synthetic data.}
We place the projective camera so that the pelvis joint is aligned with the image center and relative body size in the screen space remains unchanged. Since our silhouette synthesis network takes an unconstrained silhouette as input and generate a new silhouette in predefined view points, we separate the data generation for the source silhouettes and the target silhouettes. We render our data images at the resolution of $256\times256$. For the source silhouettes a yaw angle is randomly sampled from $360\degree$ and a pitch angle between $-10\degree$ and $60\degree$, whereas for the target silhouettes, a yaw angle is sampled from every $7.5\degree$ and a pitch angle from $10, 15, 30, 45, 60\degree$. The camera has a randomly sampled $35$mm film equivalent focal length ranged between $40$ and $135$mm for the source silhouettes and a fixed focal length of $800$mm for the target silhouettes. For the front-to-back image synthesis, we set the yaw angle to be frontal and sample the pitch angle from $0, 7.5, 15\degree$ with a focal length of $800$mm. Given the camera projection, we project $13$ joint locations that are compatible with MPII~\cite{andriluka14cvpr} onto each view point.

\begin{figure}[t]
\centering
\includegraphics[width=\linewidth]{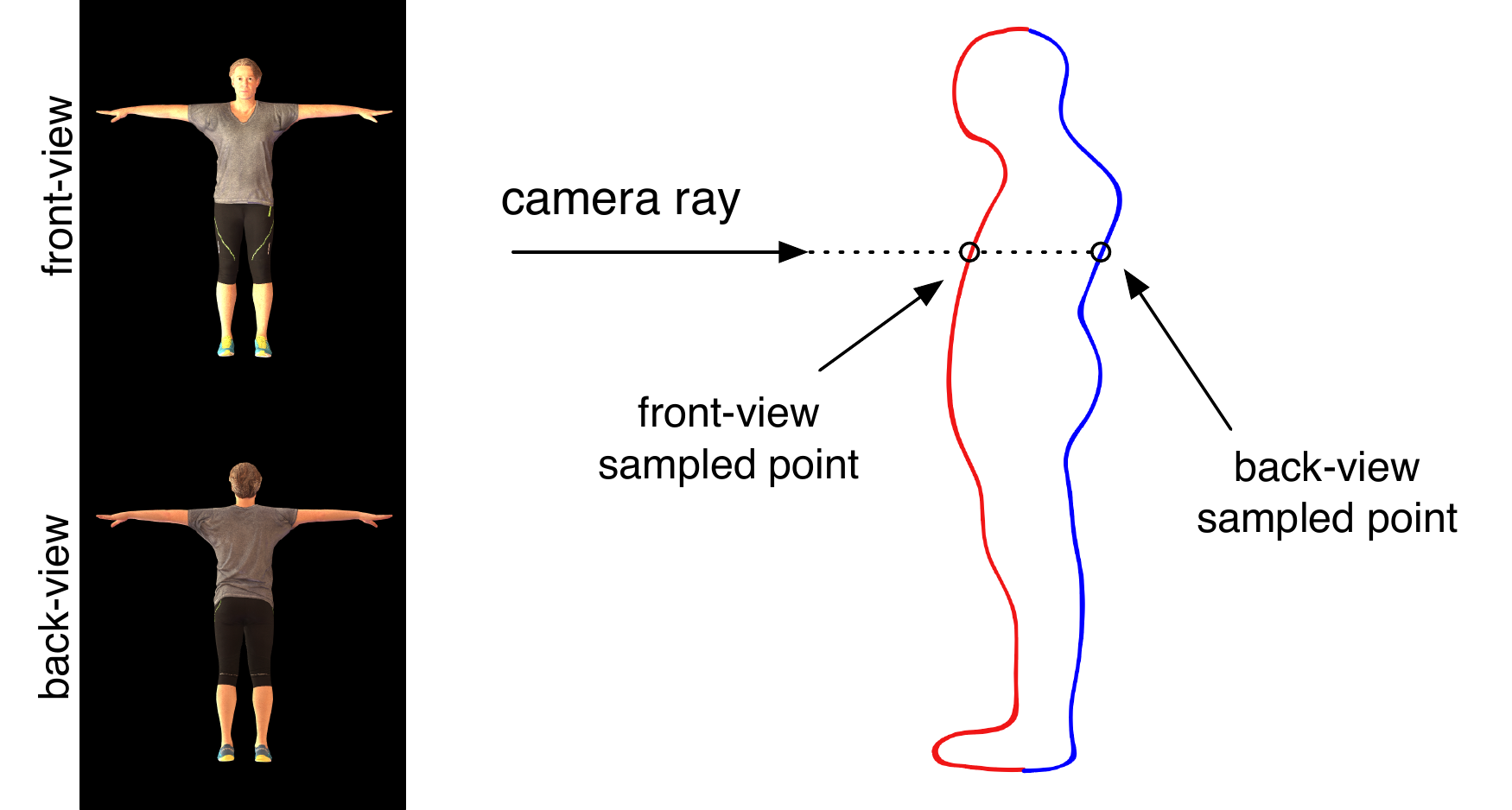}
\vspace{-15pt} 
\caption{Illustration of our back-view rendering approach.}
\label{fig:f2b_render}
\end{figure}

\paragraph{Front-to-back rendering.}
Figure~\ref{fig:f2b_render} illustrates how we generate a pair of front and back view images. Given a camera ray, normal rendering of 3D mesh sorts the depth of triangles per pixel and display the rasterization results assigned from the closest triangle. To obtain the corresponding image from the other side, we instead takes that of the furthest triangle.
Note that most common graphics libraries (e.g., OpenGL, DirectX) support this function, allowing us to generate training samples within a reasonable amount of time.


\paragraph{Network architectures.}
Both our silhouette synthesis network and the front-to-back synthesis network follow the U-Net network architecture in~\cite{pix2pix2017,yamaguchi2018high,huynh2018mesoscopic,wang2019gif2video,wang2018video} with an input channel size of $7$ and $4$, respectively. 
All the weights in these networks are initialized based on Gaussian distribution.
We use the Adam optimizer with learning rates of $2.0\times10^{-4}$, $1.0\times10^{-4}$, and $2.0\times10^{-4}$, batch size of $30$, $1$, and $1$, the number of iterations of $250,000$, $160,000$, and $50,000$, and no weight decay for the silhouette synthesis, deep visual hull, and front-to-back synthesis, respectively.
The deep visual hull is trained with the output of our silhouette synthesis network so that the distribution gap between the output of silhouette synthesis and the input for the deep visual hull algorithm is minimized. 

\paragraph{Additional networks.}
Although 2D silhouette segmentation and 3D pose estimation are not our major contributions and in practice one can use any existing methods, we train two additional networks to automatically process the input image with consistent segmentation and joint configurations.
For the silhouette segmentation, we adopt a stacked hourglass network~\cite{newell2016stacked} with three stacks.
Given an input image of resolution $256\times256\times3$, the network predicts a probability map of resolution $64\times64\times1$ for silhouettes. We further apply a deconvolution layer with a kernel size of $4$ to obtain sharper silhouettes, after concatenating $2\times$ upsampled probability and the latent features after the first convolution in the hourglass network.
The network is trained with the mean-squared error between the predicted probability map and the ground truth of UP dataset~\cite{lassner2017unite}.
For 3D pose estimation, we adopt a state-of-the-art 3D face alignment network~\cite{bulat2017far} without modification. We train the pose estimation network using our synthetically rendered body images of resolution $256\times256$ together with the corresponding 3D joints.
We use the RMSProp optimizer with a learning rate of $2.0\times10^{-5}$, a batch size of $8$ and no weight decay for training both the silhouette segmentation and pose estimation networks.


\section{Experimental Results}
\label{sec:result}

\begin{figure}[t]
\centering
\includegraphics[width=\linewidth]{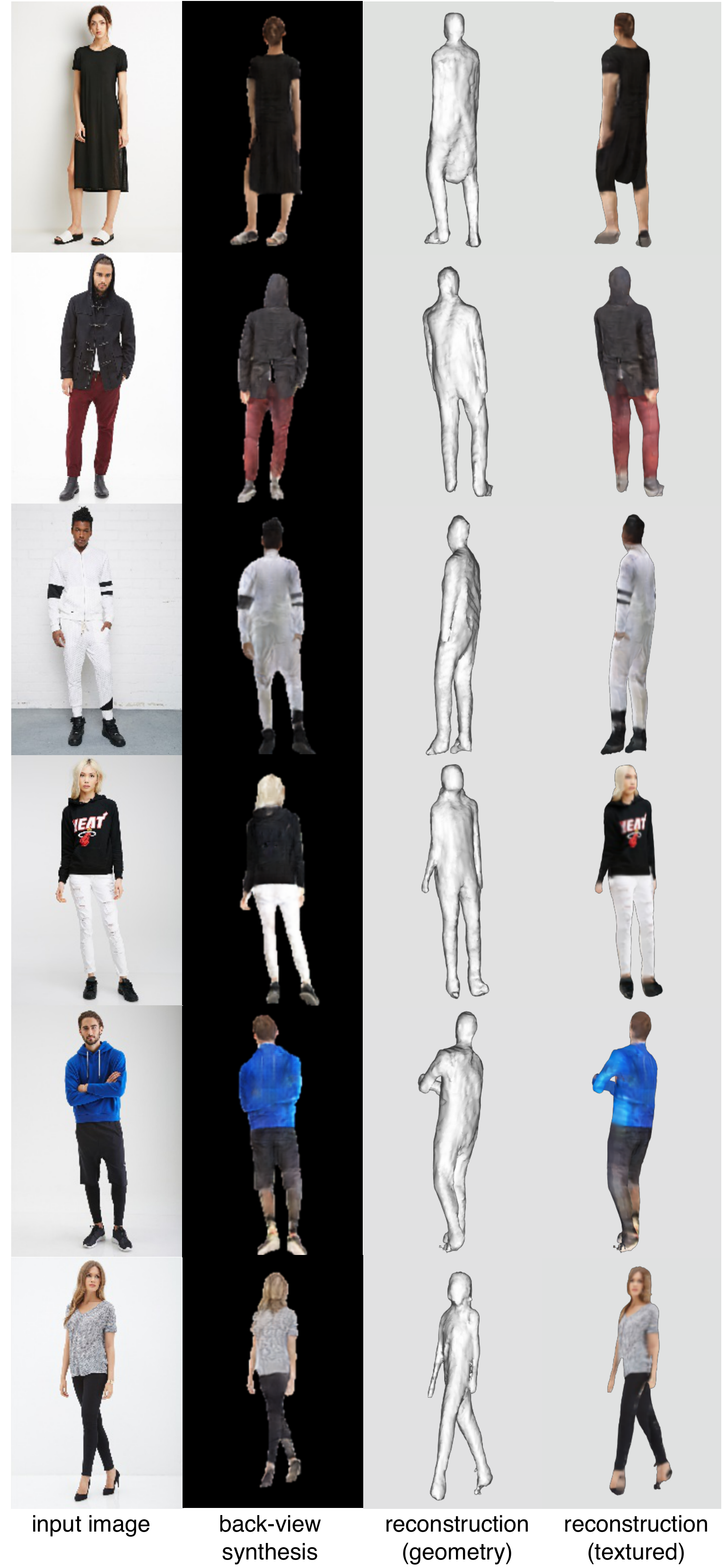}
\vspace{-12pt} 
\caption{Our 3D reconstruction results of clothed human body using test images from the DeepFashion dataset~\cite{liu2016deepfashion}.}
\label{fig:result}
\end{figure}

Figure~\ref{fig:result} shows our reconstruction results of 3D clothed human bodies with full textures on different single-view input images from the DeepFashion dataset~\cite{liu2016deepfashion}.
For each input, we show the back-view texture synthesis result, the reconstructed 3D geometry rendered with plain shading, as well as the final textured geometry.
Our method can robustly handle a variety of realistic test photos of different poses, body shapes, and cloth styles, although we train the networks using synthetically rendered images only.


\subsection{Evaluations}
\label{sec:evaluation}

\nothing{
Illustrate effectiveness of the method and superiority over existing methods:
\begin{itemize}
\item Add figure: (input photo, input texture, base geometry, output displacement map, high-frequency mesh (only after translation), super-resolution mesh), vary subjects, expressions, and zooms
\item Show better results on other capture systems (light disney one, or standard multi-view capture system, show on lee perry smith data)
\end{itemize}
}

\begin{figure}[t]
\centering
\includegraphics[width=\linewidth]{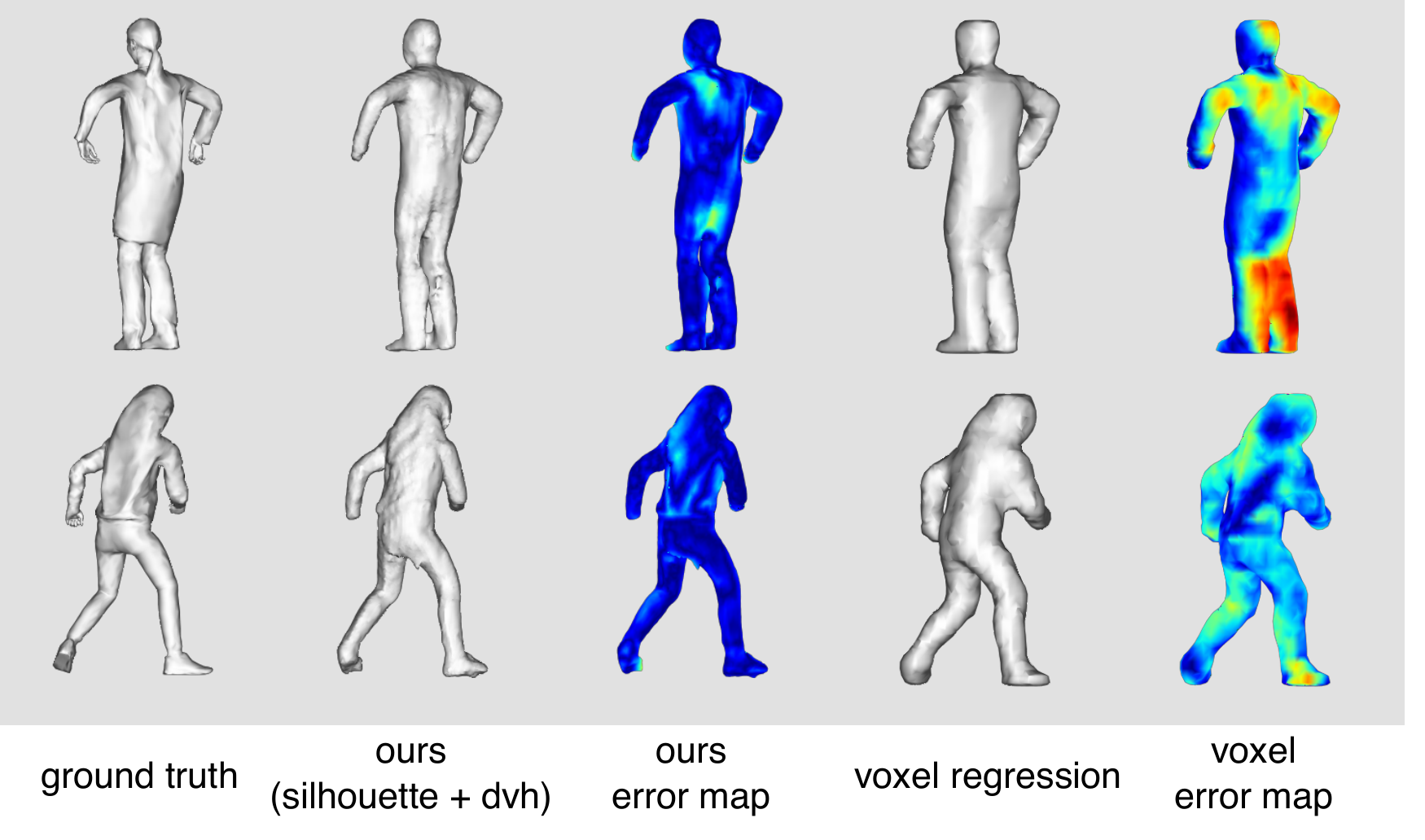}
\vspace{-15pt}
\caption{Qualitative evaluation of our silhouette-based shape representation as compared to direct voxel prediction.}
\label{fig:eval_silnet}
\end{figure}

\begin{table}

\centering
\resizebox{\columnwidth}{!}{
\begin{tabular}{ccccc}
\toprule
Input & Output & IoU (2D) & CD & EMD \\
\hline\hline
RGB+2D Pose & Silhouette & $0.826$ & $1.66$  & $4.38$  \\
Silhouette+2D Pose& Silhouette & $\mathbf{0.886}$ & $\mathbf{1.36}$  & $\mathbf{3.69}$ \\
\hline
RGB+3D Pose & Voxel & $0.471$ & $2.49$  & $5.67$ \\
Silhouette+3D Pose& Voxel & $0.462$ & $2.77$  & $6.23$ \\
\bottomrule
\end{tabular}
}
\caption{Evaluation of our silhouette-based representation compared to direct voxel prediction. The errors are measured using Chamfer Distance(CD) and Earth Mover's Distance(EDM) between the reconstructed meshes and the ground-truth.}
\label{table:eval_silnet}
\end{table}

\begin{figure}[t]
\centering
\includegraphics[width=\linewidth]{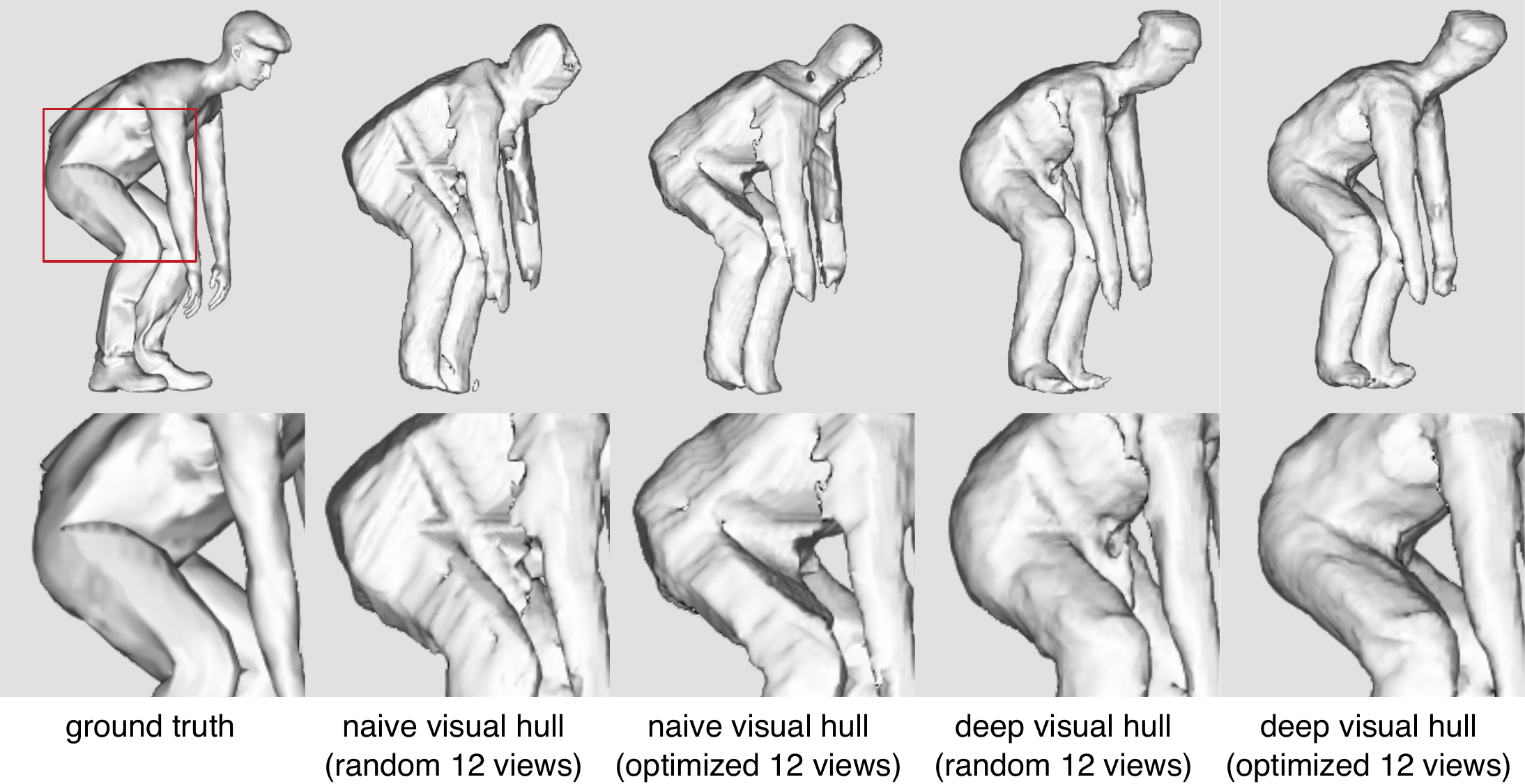}
\vspace{-7pt}
\caption{Comparisons between our deep visual hull method with a native visual hull algorithm, using both random view selection and our greedy view sampling strategy.}
\label{fig:eval_dvhull}
\end{figure}

\begin{table}
\centering
\resizebox{\columnwidth}{!}{
\begin{tabular}{ccccc}
\toprule
Input & Method & CD & EMD \\
\hline\hline
\multirow{4}{*}{\shortstack{Inferred \\ silhouettes}}  & visual hull (random) & $2.12$ & $6.95$ \\
 & visual hull (optimized) & $1.37$ & $6.93$ \\
 & deep v-hull (random)  & $1.41$  & $3.79$ \\
 & deep v-hull (optimized) & $\mathbf{1.34}$  & $\mathbf{3.66}$  \\
\hline
GT silhouettes & visual hull (8 views) & $0.67$  & $3.19$ \\
GT images &  Huang et al.~\cite{Huang18ECCV} (4 views)  & $0.98$  & $4.09$  \\
\bottomrule
\end{tabular}
}
\caption{Evaluation of our greedy sampling method to compute deep visual hull.
}
\label{table:eval_vhull}
\end{table}

\paragraph{Silhouette Representation.}
We verify the effectiveness of our silhouette-based representation by comparing it with several alternative approaches based on the Renderpeople dataset, including direct voxel prediction from 3D pose and using RGB image to replace 2D silhouette as input for deep visual hull algorithm.
Please refer to our supplementary materials for implementation details of our baseline methods used for comparisons.
For all the methods, we report (1) the 2D Intersection over Union (IoU) for the synthetically generated side view and (2) the 3D reconstruction error based on Chamfer distances between the reconstructed meshes and the ground-truth (in centimeter) in Table~\ref{table:eval_silnet}.
It is evident that direct voxel prediction will lead to poor accuracy when matching the side view in 2D and aligning with the ground-truth geometry in 3D, as compared to our silhouette-based representation.
Figure~\ref{fig:eval_silnet} shows qualitative comparisons demonstrating the advantages of our silhouette-based representation.


\paragraph{Visual hull reconstruction.}
In Table~\ref{table:eval_vhull} and Figure~\ref{fig:eval_dvhull}, we compare our deep visual hull algorithm (Sec.~\ref{sec:dvhull}) with a naive visual hull method.
We also evaluate our greedy view sampling strategy by comparing it with random view selection. We use $12$ inferred silhouettes as input for different methods and evaluate the reconstruction errors using Chamfer distances.
For random view selection, we repeat the process $100$ times and compute the average error.
As additional references, we also provide the corresponding results using the naive visual hull method with $8$ ground-truth silhouettes, as well as the method in~\cite{Huang18ECCV} using $4$ ground-truth images.
As shown in Table~\ref{table:eval_vhull}, our deep visual algorithm outperforms a naive approach and our greedy view sampling strategy can significantly improve the results in terms of reconstruction errors.
In addition, for the deep visual hull algorithm, our view sampling strategy is better than $69\%$ random selected views, while for the naive visual hull method, our approach always outperforms random view selection.
Figure~\ref{fig:eval_dvhull} demonstrates that our deep visual hull method helps fix some artifacts and missing parts especially in concave regions, which are caused by inconsistency among multi-view silhouettes synthesis results.
%


\subsection{Comparisons}
\label{sec:comparisons}

\begin{figure}[t]
\centering
\includegraphics[width=\linewidth]{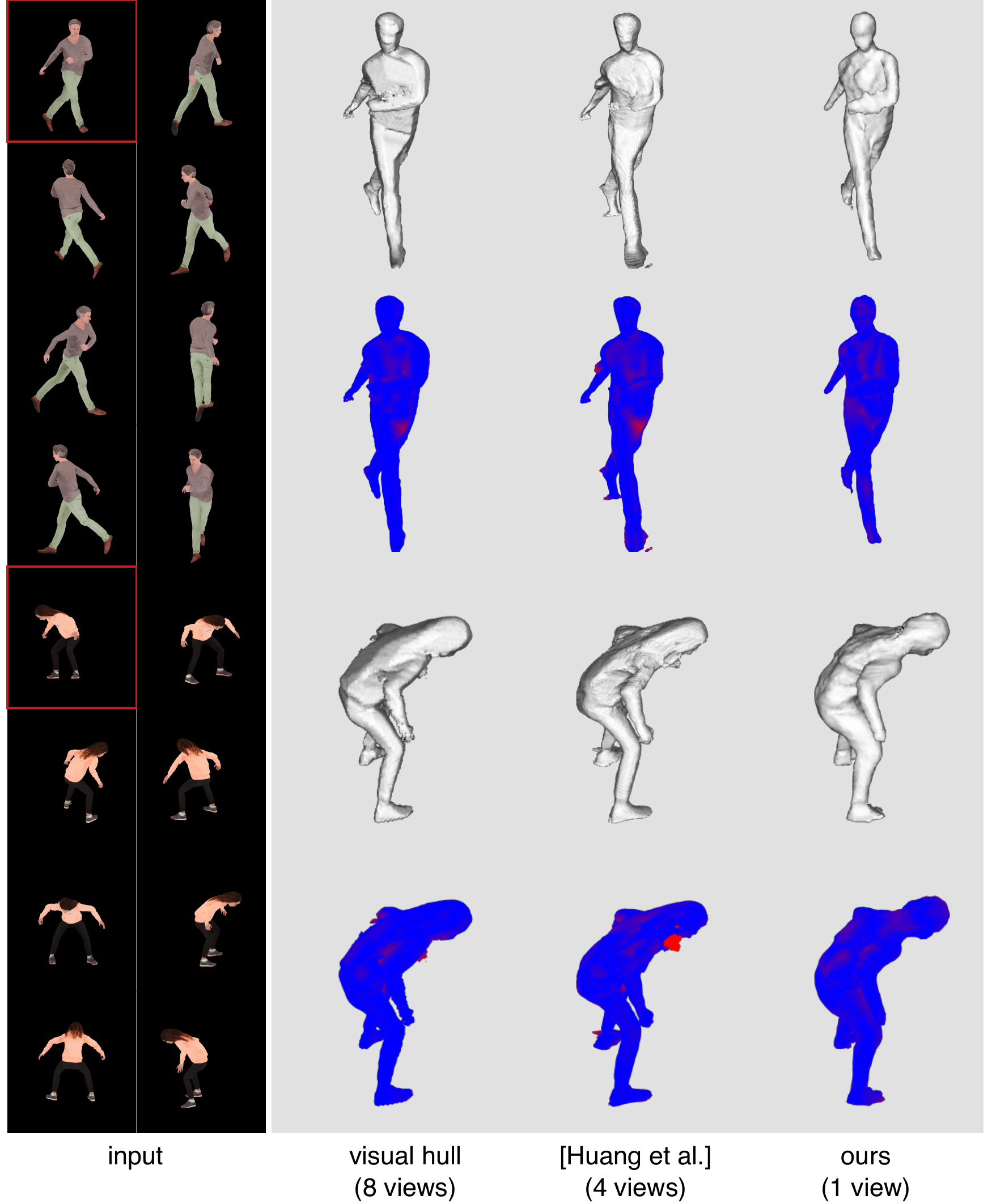}
\vspace{-7pt}
\caption{Comparison with multiview visual hull algorithms. Despite the single view input, our method produces comparable reconstruction results. Note that input image in red color is the single-view input for our method and the top four views are used for Huang~\etal~\cite{Huang18ECCV}.
}
\label{fig:comp_vhull}
\end{figure}

In Figure~\ref{fig:comp_vhull}, we compare our method using single-view input with a native visual hull algorithm using $8$ input views as well as Huang et al.~\cite{Huang18ECCV} using four input views.
For each result, we show both the plain shaded 3D geometry and the color-coded 3D reconstruction error.
Although using a single image as input each time, we can still generate results that are visually comparable to those from methods based on multi-view input.

\begin{figure}[t]
\centering
\includegraphics[width=\linewidth, trim=55px 0px 55px 0px,
  clip]{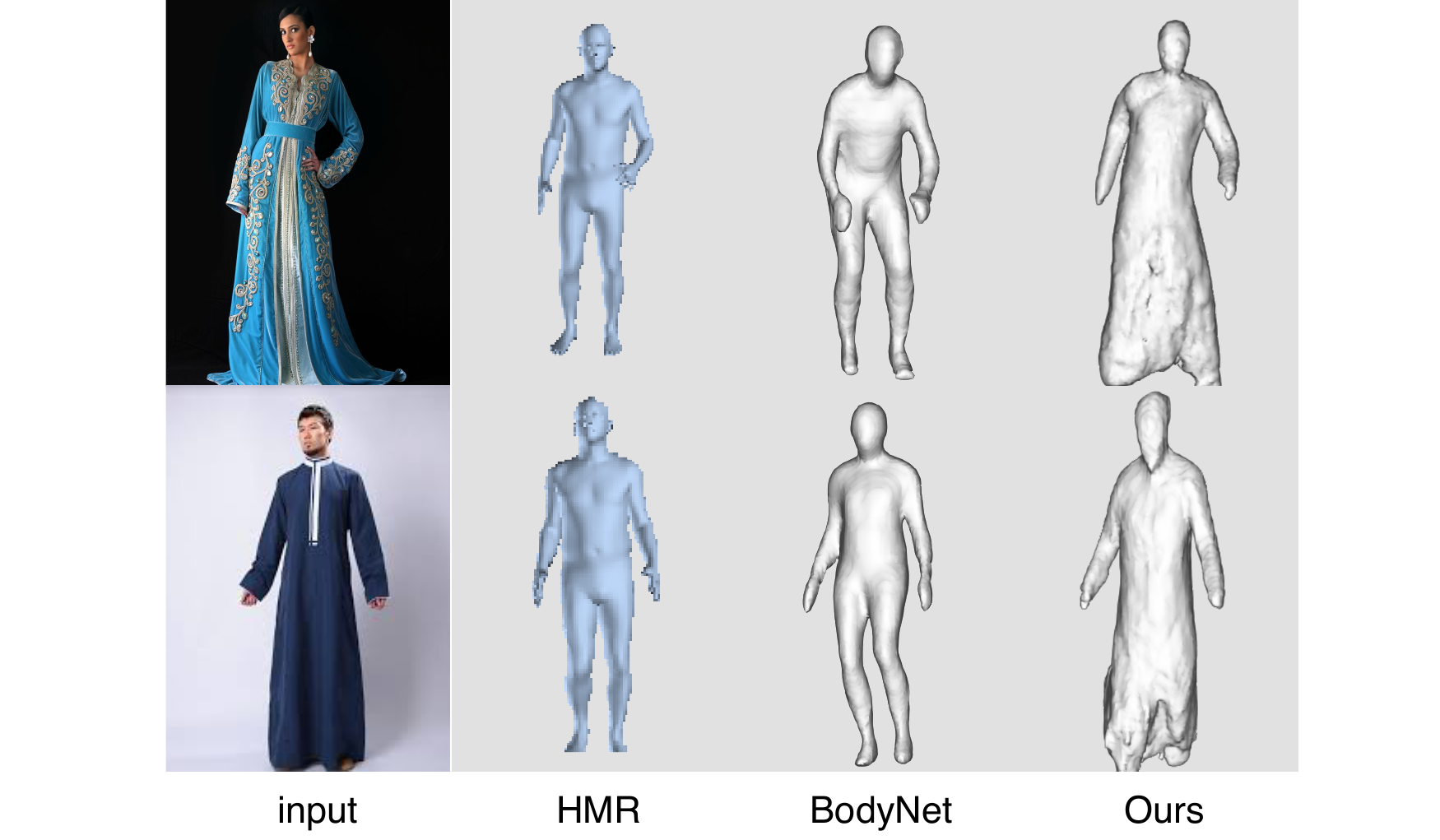}
\vspace{-7pt} 
\caption{We qualitatively compare our method with two state-of-the-art single view human reconstruction techniques, HMR~\cite{hmrKanazawa17} and BodyNet~\cite{varol18_bodynet}.
}
\label{fig:comp_singleview}
\end{figure}

In Figure~\ref{fig:comp_singleview}, we qualitatively compare our results with state-of-the-art single-view human reconstruction techniques~\cite{hmrKanazawa17,varol18_bodynet}.
Since existing methods focus on body shape only using parametric models, our approach can generate more faithful results in cases of complex clothed geometry.

\nothing{
\begin{itemize}
    \item comparison with HMR \cite{hmrKanazawa17} and BodyNet \cite{varol18_bodynet}
    \item comparison with zeng's eccv \cite{Huang18ECCV} and visual hull with more views
\end{itemize}
}

\section{Discussion and Future Work}
\label{sec:conclusion}

\begin{figure}[t]
\begin{center}
 \includegraphics[width=\linewidth, trim=15px 0px 15px 0px,
  clip]{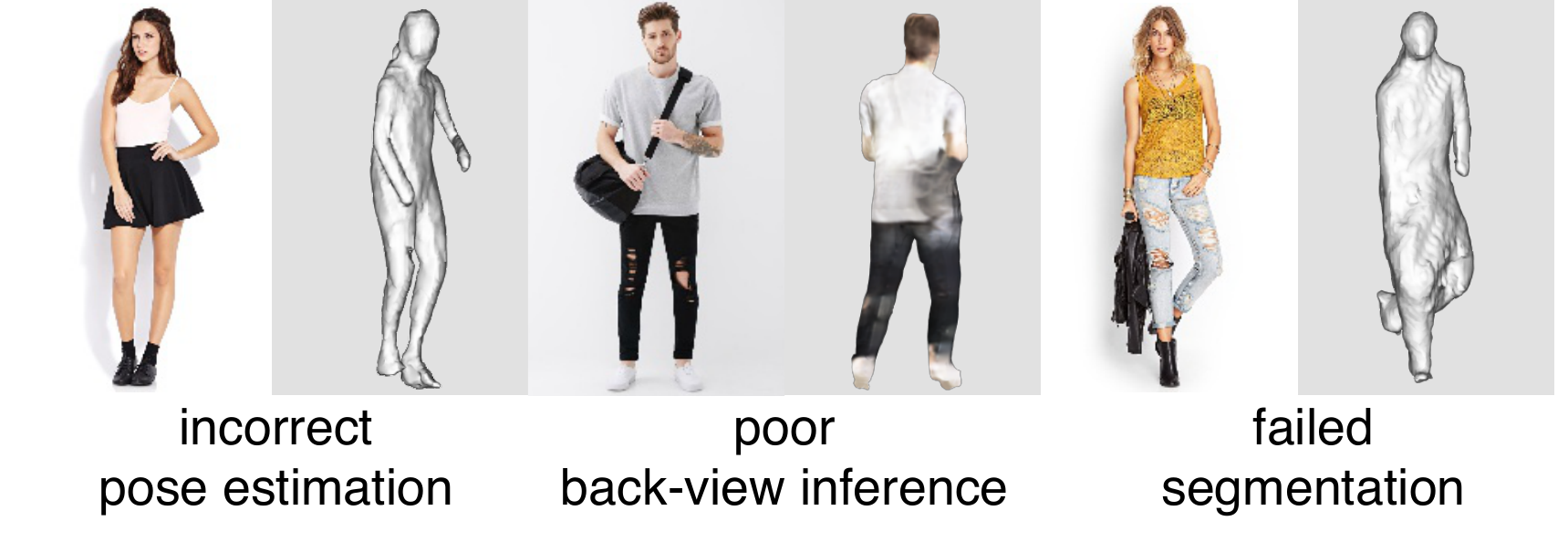}
\end{center}
\vspace{-15pt} 
\caption{Failure cases.}
\label{fig:failure}
\end{figure}

In this paper, we present a framework for monocular 3D human reconstruction using deep neural networks.
From a single input image of the subject, we can predict the 3D textured geometry of clothed body shape, without any requirement of a parametric model or a pre-captured template.
To this end, we propose a novel-view silhouette synthesis network based on adversarial training, an improved deep visual hull algorithm with a greedy view selection strategy, as well as a front-to-back texture synthesis network.

One major limitation of our current implementation is that our synthetic training data is very limited and may be biased from real images.
See Figure~\ref{fig:failure} for a few typical failure cases, in which the 3D pose estimation may fail or there are some additional accessories not covered by our training data.
It would be helpful to add realistic training data which may be tedious and costly to acquire.
The output mesh using our method is not rigged and thus cannot be directly used for animation.
Also we do not explicitly separate the geometry of cloth and human body.
In the future, we plan to extend our method to predict output with high frequency details and semantic labels.
Finally, it is interesting to infer relightable textures such as diffuse and specular albedo maps.


\section*{\small{Acknowledgements}}
\small{
Shigeo Morishima is supported by the JST ACCEL Grant Number JPMJAC1602, JSPS KAKENHI Grant Number JP17H06101, the Waseda Research Institute for Science and Engineering. Hao Li is affiliated with the University of Southern California, the USC Institute for Creative Technologies, and Pinscreen. This research was conducted at USC and was funded by in part by the ONR YIP grant N00014-17-S-FO14, the CONIX Research Center, one of six centers in JUMP, a Semiconductor Research Corporation (SRC) program sponsored by DARPA, the Andrew and Erna Viterbi Early Career Chair, the U.S. Army Research Laboratory (ARL) under contract number W911NF-14-D-0005, Adobe, and Sony. This project was not funded by Pinscreen, nor has it been conducted at Pinscreen or by anyone else affiliated with Pinscreen. The content of the information does not necessarily reflect the position or the policy of the Government, and no official endorsement should be inferred.
}

{\small
	\bibliographystyle{ieee}
	\bibliography{paper}

\begin{thebibliography}{10}\itemsep=-1pt

\bibitem{alldieck2018detailed}
T.~Alldieck, M.~Magnor, W.~Xu, C.~Theobalt, and G.~Pons-Moll.
\newblock Detailed human avatars from monocular video.
\newblock In {\em International Conference on 3D Vision}, pages 98--109, 2018.

\bibitem{alldieck2018video}
T.~Alldieck, M.~A. Magnor, W.~Xu, C.~Theobalt, and G.~Pons-Moll.
\newblock Video based reconstruction of 3d people models.
\newblock In {\em IEEE Conference on Computer Vision and Pattern Recognition},
  pages 8387--8397, 2018.

\bibitem{anguelov2005scape}
D.~Anguelov, P.~Srinivasan, D.~Koller, S.~Thrun, J.~Rodgers, and J.~Davis.
\newblock {SCAPE: shape completion and animation of people}.
\newblock {\em ACM Transactions on Graphics}, 24(3):408--416, 2005.

\bibitem{balakrishnan2018synthesizing}
G.~Balakrishnan, A.~Zhao, A.~V. Dalca, F.~Durand, and J.~Guttag.
\newblock Synthesizing images of humans in unseen poses.
\newblock In {\em IEEE Conference on Computer Vision and Pattern Recognition},
  pages 8340--8348, 2018.

\bibitem{balan2007detailed}
A.~O. Balan, L.~Sigal, M.~J. Black, J.~E. Davis, and H.~W. Haussecker.
\newblock Detailed human shape and pose from images.
\newblock In {\em IEEE Conference on Computer Vision and Pattern Recognition},
  pages 1--8, 2007.

\bibitem{bogo2016keep}
F.~Bogo, A.~Kanazawa, C.~Lassner, P.~Gehler, J.~Romero, and M.~J. Black.
\newblock {Keep it SMPL: Automatic estimation of 3D human pose and shape from a
  single image}.
\newblock In {\em European Conference on Computer Vision}, pages 561--578,
  2016.

\bibitem{bulat2017far}
A.~Bulat and G.~Tzimiropoulos.
\newblock How far are we from solving the 2d \& 3d face alignment problem?(and
  a dataset of 230,000 3d facial landmarks).
\newblock In {\em IEEE International Conference on Computer Vision}, pages
  1021--1030, 2017.

\bibitem{cao2017realtime}
Z.~Cao, T.~Simon, S.-E. Wei, and Y.~Sheikh.
\newblock Realtime multi-person {2D} pose estimation using part affinity
  fields.
\newblock In {\em IEEE Conference on Computer Vision and Pattern Recognition},
  pages 7291--7299, 2017.

\bibitem{cheung2003visual}
G.~K. Cheung, S.~Baker, and T.~Kanade.
\newblock Visual hull alignment and refinement across time: A {3D}
  reconstruction algorithm combining shape-from-silhouette with stereo.
\newblock In {\em IEEE Conference on Computer Vision and Pattern Recognition},
  pages 375--382, 2003.

\bibitem{dibra2016hs}
E.~Dibra, H.~Jain, C.~{\"O}ztireli, R.~Ziegler, and M.~Gross.
\newblock Hs-nets: Estimating human body shape from silhouettes with
  convolutional neural networks.
\newblock In {\em International Conference on 3D Vision}, pages 108--117, 2016.

\bibitem{dibra2017human}
E.~Dibra, H.~Jain, C.~Oztireli, R.~Ziegler, and M.~Gross.
\newblock Human shape from silhouettes using generative hks descriptors and
  cross-modal neural networks.
\newblock In {\em IEEE Conference on Computer Vision and Pattern Recognition},
  pages 4826--4836, 2017.

\bibitem{esser2018variational}
P.~Esser, E.~Sutter, and B.~Ommer.
\newblock A variational u-net for conditional appearance and shape generation.
\newblock In {\em IEEE Conference on Computer Vision and Pattern Recognition},
  pages 8857--8866, 2018.

\bibitem{esteban2004silhouette}
C.~H. Esteban and F.~Schmitt.
\newblock {Silhouette and stereo fusion for 3D object modeling}.
\newblock {\em Computer Vision and Image Understanding}, 96(3):367--392, 2004.

\bibitem{franco2006visual}
J.-S. Franco, M.~Lapierre, and E.~Boyer.
\newblock Visual shapes of silhouette sets.
\newblock In {\em International Symposium on 3D Data Processing, Visualization,
  and Transmission}, pages 397--404, 2006.

\bibitem{furukawa2006carved}
Y.~Furukawa and J.~Ponce.
\newblock Carved visual hulls for image-based modeling.
\newblock In {\em European Conference on Computer Vision}, pages 564--577,
  2006.

\bibitem{furukawa2010accurate}
Y.~Furukawa and J.~Ponce.
\newblock Accurate, dense, and robust multiview stereopsis.
\newblock {\em IEEE Transactions on Pattern Analysis and Machine Intelligence},
  32(8):1362--1376, 2010.

\bibitem{gilbert:eccv:2018}
A.~Gilbert, M.~Volino, J.~Collomosse, and A.~Hilton.
\newblock Volumetric performance capture from minimal camera viewpoints.
\newblock In {\em European Conference on Computer Vision}, pages 566--581,
  2018.

\bibitem{guan2009estimating}
P.~Guan, A.~Weiss, A.~O. Balan, and M.~J. Black.
\newblock Estimating human shape and pose from a single image.
\newblock In {\em IEEE International Conference on Computer Vision}, pages
  1381--1388, 2009.

\bibitem{guler2018densepose}
R.~A. G{\"u}ler, N.~Neverova, and I.~Kokkinos.
\newblock Densepose: Dense human pose estimation in the wild.
\newblock In {\em IEEE Conference on Computer Vision and Pattern Recognition},
  pages 7297--7306, 2018.

\bibitem{Huang18ECCV}
Z.~Huang, T.~Li, W.~Chen, Y.~Zhao, J.~Xing, C.~LeGendre, L.~Luo, C.~Ma, and
  H.~Li.
\newblock Deep volumetric video from very sparse multi-view performance
  capture.
\newblock In {\em European Conference on Computer Vision}, pages 336--354,
  2018.

\bibitem{huynh2018mesoscopic}
L.~Huynh, W.~Chen, S.~Saito, J.~Xing, K.~Nagano, A.~Jones, P.~Debevec, and
  H.~Li.
\newblock Mesoscopic facial geometry inference using deep neural networks.
\newblock In {\em Proceedings of the IEEE Conference on Computer Vision and
  Pattern Recognition}, pages 8407--8416, 2018.

\bibitem{pix2pix2017}
P.~Isola, J.-Y. Zhu, T.~Zhou, and A.~A. Efros.
\newblock Image-to-image translation with conditional adversarial networks.
\newblock In {\em IEEE Conference on Computer Vision and Pattern Recognition},
  pages 1125--1134, 2017.

\bibitem{hmrKanazawa17}
A.~Kanazawa, M.~J. Black, D.~W. Jacobs, and J.~Malik.
\newblock End-to-end recovery of human shape and pose.
\newblock In {\em IEEE Conference on Computer Vision and Pattern Recognition},
  pages 7122--7131, 2018.

\bibitem{lassner2017unite}
C.~Lassner, J.~Romero, M.~Kiefel, F.~Bogo, M.~J. Black, and P.~V. Gehler.
\newblock {Unite the people: Closing the loop between 3d and 2d human
  representations}.
\newblock In {\em IEEE Conference on Computer Vision and Pattern Recognition},
  pages 6050--6059, 2017.

\bibitem{li2018context}
H.~Li, G.~Li, L.~Lin, H.~Yu, and Y.~Yu.
\newblock Context-aware semantic inpainting.
\newblock {\em IEEE Transactions on Cybernetics}, 2018.

\bibitem{li2013three}
H.~Li, E.~Vouga, A.~Gudym, L.~Luo, J.~T. Barron, and G.~Gusev.
\newblock {3D} self-portraits.
\newblock {\em ACM Transactions on Graphics}, 32(6):187:1--187:9, 2013.

\bibitem{liu2016deepfashion}
Z.~Liu, P.~Luo, S.~Qiu, X.~Wang, and X.~Tang.
\newblock Deepfashion: Powering robust clothes recognition and retrieval with
  rich annotations.
\newblock In {\em IEEE Conference on Computer Vision and Pattern Recognition},
  pages 1096--1104, 2016.

\bibitem{loper2015smpl}
M.~Loper, N.~Mahmood, J.~Romero, G.~Pons-Moll, and M.~J. Black.
\newblock {SMPL: A skinned multi-person linear model}.
\newblock {\em ACM Transactions on Graphics}, 34(6):248:1--248:16, 2015.

\bibitem{ma2017pose}
L.~Ma, X.~Jia, Q.~Sun, B.~Schiele, T.~Tuytelaars, and L.~Van~Gool.
\newblock Pose guided person image generation.
\newblock In {\em Advances in Neural Information Processing Systems}, pages
  406--416, 2017.

\bibitem{ma2018disentangled}
L.~Ma, Q.~Sun, S.~Georgoulis, L.~Van~Gool, B.~Schiele, and M.~Fritz.
\newblock Disentangled person image generation.
\newblock In {\em IEEE Conference on Computer Vision and Pattern Recognition},
  pages 99--108, 2018.

\bibitem{matusik2000image}
W.~Matusik, C.~Buehler, R.~Raskar, S.~J. Gortler, and L.~McMillan.
\newblock Image-based visual hulls.
\newblock In {\em ACM SIGGRAPH}, pages 369--374, 2000.

\bibitem{mehta2017vnect}
D.~Mehta, S.~Sridhar, O.~Sotnychenko, H.~Rhodin, M.~Shafiei, H.-P. Seidel,
  W.~Xu, D.~Casas, and C.~Theobalt.
\newblock {VNect: Real-time 3D Human Pose Estimation with a Single RGB Camera}.
\newblock {\em ACM Transactions on Graphics}, 36(4):44:1--44:14, 2017.

\bibitem{andriluka14cvpr}
A.~Mykhaylo, P.~Leonid, G.~Peter, and B.~Schiele.
\newblock 2d human pose estimation: New benchmark and state of the art
  analysis.
\newblock In {\em IEEE Conference on Computer Vision and Pattern Recognition},
  pages 3686--3693, 2014.

\bibitem{NewcombeFS15}
R.~A. Newcombe, D.~Fox, and S.~M. Seitz.
\newblock {DynamicFusion}: Reconstruction and tracking of non-rigid scenes in
  real-time.
\newblock In {\em IEEE Conference on Computer Vision and Pattern Recognition},
  pages 343--352, 2015.

\bibitem{newell2016stacked}
A.~Newell, K.~Yang, and J.~Deng.
\newblock Stacked hourglass networks for human pose estimation.
\newblock In {\em European Conference on Computer Vision}, pages 483--499,
  2016.

\bibitem{pons2017clothcap}
G.~Pons-Moll, S.~Pujades, S.~Hu, and M.~J. Black.
\newblock Clothcap: Seamless 4d clothing capture and retargeting.
\newblock {\em ACM Transactions on Graphics}, 36(4):73:1--73:15, 2017.

\bibitem{rematas2017novel}
K.~Rematas, C.~H. Nguyen, T.~Ritschel, M.~Fritz, and T.~Tuytelaars.
\newblock Novel views of objects from a single image.
\newblock {\em IEEE Transactions on Pattern Analysis and Machine Intelligence},
  39(8):1576--1590, 2017.

\bibitem{Rogez:2018:LCR-Net}
G.~Rogez, P.~Weinzaepfel, and C.~Schmid.
\newblock {LCR-Net++: Multi-person 2D and 3D Pose Detection in Natural Images}.
\newblock {\em arXiv preprint arXiv:1803.00455}, 2018.

\bibitem{Seitz:2006}
S.~M. Seitz, B.~Curless, J.~Diebel, D.~Scharstein, and R.~Szeliski.
\newblock A comparison and evaluation of multi-view stereo reconstruction
  algorithms.
\newblock In {\em IEEE Conference on Computer Vision and Pattern Recognition},
  pages 519--528, 2006.

\bibitem{simonyan2014very}
K.~Simonyan and A.~Zisserman.
\newblock Very deep convolutional networks for large-scale image recognition.
\newblock {\em arXiv preprint arXiv:1409.1556}, 2014.

\bibitem{starck2007surface}
J.~Starck and A.~Hilton.
\newblock Surface capture for performance-based animation.
\newblock {\em IEEE Computer Graphics and Applications}, 27(3):21--31, 2007.

\bibitem{tan2017indirect}
J.~Tan, I.~Budvytis, and R.~Cipolla.
\newblock Indirect deep structured learning for 3d human body shape and pose
  prediction.
\newblock In {\em British Machine Vision Conference}, pages 6.1--6.11, 2017.

\bibitem{tung2017self}
H.-Y. Tung, H.-W. Tung, E.~Yumer, and K.~Fragkiadaki.
\newblock Self-supervised learning of motion capture.
\newblock In {\em Advances in Neural Information Processing Systems}, pages
  5236--5246, 2017.

\bibitem{varol18_bodynet}
G.~Varol, D.~Ceylan, B.~Russell, J.~Yang, E.~Yumer, I.~Laptev, and C.~Schmid.
\newblock {BodyNet}: Volumetric inference of {3D} human body shapes.
\newblock In {\em European Conference on Computer Vision}, pages 20--36, 2018.

\bibitem{vlasic2008articulated}
D.~Vlasic, I.~Baran, W.~Matusik, and J.~Popovi{\'c}.
\newblock Articulated mesh animation from multi-view silhouettes.
\newblock {\em ACM Transactions on Graphics}, 27(3):97:1--97:9, 2008.

\bibitem{vlasic2009dynamic}
D.~Vlasic, P.~Peers, I.~Baran, P.~Debevec, J.~Popovi{\'c}, S.~Rusinkiewicz, and
  W.~Matusik.
\newblock Dynamic shape capture using multi-view photometric stereo.
\newblock {\em ACM Transactions on Graphics}, 28(5):174:1--174:11, 2009.

\bibitem{wang2018video}
C.~Wang, H.~Huang, X.~Han, and J.~Wang.
\newblock Video inpainting by jointly learning temporal structure and spatial
  details.
\newblock {\em arXiv preprint arXiv:1806.08482}, 2018.

\bibitem{wang2018pix2pixHD}
T.-C. Wang, M.-Y. Liu, J.-Y. Zhu, A.~Tao, J.~Kautz, and B.~Catanzaro.
\newblock High-resolution image synthesis and semantic manipulation with
  conditional gans.
\newblock In {\em IEEE Conference on Computer Vision and Pattern Recognition},
  pages 8798--8807, 2018.

\bibitem{wang2019gif2video}
Y.~Wang, H.~Huang, C.~Wang, T.~He, J.~Wang, and M.~Hoai.
\newblock Gif2video: Color dequantization and temporal interpolation of gif
  images.
\newblock {\em arXiv preprint arXiv:1901.02840}, 2019.

\bibitem{waschbusch2005scalable}
M.~Waschb{\"u}sch, S.~W{\"u}rmlin, D.~Cotting, F.~Sadlo, and M.~Gross.
\newblock Scalable {3D} video of dynamic scenes.
\newblock {\em The Visual Computer}, 21(8):629--638, 2005.

\bibitem{wei2016cpm}
S.-E. Wei, V.~Ramakrishna, T.~Kanade, and Y.~Sheikh.
\newblock Convolutional pose machines.
\newblock In {\em IEEE Conference on Computer Vision and Pattern Recognition},
  pages 4724--4732, 2016.

\bibitem{weng2018photo}
C.-Y. Weng, B.~Curless, and I.~Kemelmacher-Shlizerman.
\newblock Photo wake-up: 3d character animation from a single photo.
\newblock {\em arXiv preprint arXiv:1812.02246}, 2018.

\bibitem{wu2011shading}
C.~Wu, K.~Varanasi, Y.~Liu, H.-P. Seidel, and C.~Theobalt.
\newblock Shading-based dynamic shape refinement from multi-view video under
  general illumination.
\newblock In {\em IEEE International Conference on Computer Vision}, pages
  1108--1115, 2011.

\bibitem{Xu:2018:MHP}
W.~Xu, A.~Chatterjee, M.~Zollh\"{o}fer, H.~Rhodin, D.~Mehta, H.-P. Seidel, and
  C.~Theobalt.
\newblock Monoperfcap: Human performance capture from monocular video.
\newblock {\em ACM Transactions on Graphics}, 37(2):27:1--27:15, 2018.

\bibitem{yamaguchi2018high}
S.~Yamaguchi, S.~Saito, K.~Nagano, Y.~Zhao, W.~Chen, K.~Olszewski,
  S.~Morishima, and H.~Li.
\newblock High-fidelity facial reflectance and geometry inference from an
  unconstrained image.
\newblock {\em ACM Transactions on Graphics}, 37(4):162, 2018.

\bibitem{yang2016estimation}
J.~Yang, J.-S. Franco, F.~H{\'e}troy-Wheeler, and S.~Wuhrer.
\newblock Estimation of human body shape in motion with wide clothing.
\newblock In {\em European Conference on Computer Vision}, pages 439--454,
  2016.

\bibitem{yang20183d}
W.~Yang, W.~Ouyang, X.~Wang, J.~Ren, H.~Li, and X.~Wang.
\newblock {3D} human pose estimation in the wild by adversarial learning.
\newblock In {\em IEEE Conference on Computer Vision and Pattern Recognition},
  pages 5255--5264, 2018.

\bibitem{zhang-CVPR17}
C.~Zhang, S.~Pujades, M.~Black, and G.~Pons-Moll.
\newblock Detailed, accurate, human shape estimation from clothed {3D} scan
  sequences.
\newblock In {\em IEEE Conference on Computer Vision and Pattern Recognition},
  pages 4191--4200, 2017.

\bibitem{zhang2018densely}
H.~Zhang and V.~M. Patel.
\newblock Densely connected pyramid dehazing network.
\newblock In {\em IEEE Conference on Computer Vision and Pattern Recognition},
  pages 3194--3203, 2018.

\bibitem{zhang2017image}
H.~Zhang, V.~Sindagi, and V.~M. Patel.
\newblock Image de-raining using a conditional generative adversarial network.
\newblock {\em arXiv preprint arXiv:1701.05957}, 2017.

\bibitem{zhou2016view}
T.~Zhou, S.~Tulsiani, W.~Sun, J.~Malik, and A.~A. Efros.
\newblock View synthesis by appearance flow.
\newblock In {\em European conference on computer vision}, pages 286--301,
  2016.

\bibitem{zhu2018view}
H.~Zhu, H.~Su, P.~Wang, X.~Cao, and R.~Yang.
\newblock View extrapolation of human body from a single image.
\newblock In {\em IEEE Conference on Computer Vision and Pattern Recognition},
  pages 4450--4459, 2018.

\bibitem{zitnick2004high}
C.~L. Zitnick, S.~B. Kang, M.~Uyttendaele, S.~Winder, and R.~Szeliski.
\newblock High-quality video view interpolation using a layered representation.
\newblock {\em ACM Transactions on Graphics}, 23(3):600--608, 2004.

\end{thebibliography}
}

\ifthenelse{\equal{\final}{0}}
{
}
{}
\end{document}